\pdfoutput=1

\documentclass[11pt]{article}

\usepackage[]{acl}

\usepackage{times}
\usepackage{latexsym}
\usepackage{amssymb}

\usepackage[T1]{fontenc}

\usepackage[utf8]{inputenc}

\usepackage{microtype}

\usepackage{amsmath}
\usepackage{amssymb}
\usepackage{amsfonts}
\usepackage{graphicx}
\usepackage[toc,page]{appendix}
\usepackage{geometry}
\usepackage{marvosym}
\usepackage{xltabular}
\usepackage{titling}
\usepackage{xcolor}
\usepackage{hyperref}
\usepackage{latexsym}
\usepackage{subfigure}

\title{From Understanding to Utilization: A Survey on Explainability for Large Language Models}

\author{Haoyan Luo \\
  Imperial College London \\
  \texttt{h.luo23@imperial.ac.uk} \\
  \And
  Lucia Specia \\
  Imperial College London \\
  \texttt{l.specia@imperial.ac.uk} \\
  \\}

\begin{document}
\maketitle
\begin{abstract}
    Explainability for Large Language Models (LLMs) is a critical yet challenging aspect of natural language processing. As LLMs are increasingly integral to diverse applications, their ``black-box'' nature sparks significant concerns regarding transparency and ethical use. 
    This survey underscores the imperative for increased explainability in LLMs, delving into both the research on explainability and the various methodologies and tasks that utilize an understanding of these models. 
    Our focus is primarily on pre-trained Transformer-based LLMs, such as LLaMA \cite{touvron2023llama}, which pose distinctive interpretability challenges due to their scale and complexity.
    In terms of existing methods, we classify them into local and global analyses, based on their explanatory objectives. 
    When considering the utilization of explainability, we explore several compelling methods that concentrate on model editing, control generation, and model enhancement.
    Additionally, we examine representative evaluation metrics and datasets, elucidating their advantages and limitations. 
    Our goal is to reconcile theoretical and empirical understanding with practical implementation, proposing exciting avenues for explanatory techniques and their applications in the LLMs era.
\end{abstract}

\section{Introduction}
    In the rapidly evolving field of natural language processing, Large Language Models (LLMs) have emerged as a cornerstone, demonstrating remarkable proficiency across a spectrum of tasks. Despite their effectiveness, LLMs, often characterized as ``black-box'' systems, present a substantial challenge in terms of explainability and transparency. This opacity can lead to unintended consequences, such as the generation of harmful or misleading content \cite{gehman-etal-2020-realtoxicityprompts}, and the occurrence of model hallucinations \cite{weidinger2021ethical}. These issues underscore the urgency for improved explainability, not just for understanding, but for responsible and ethical application.
    
    
    Explainability in LLMs serves two critical functions. For end users, it fosters trust by clarifying the model's reasoning in a nontechnical manner, enhancing understanding of their capabilities and potential flaws \cite{zhao2023explainability}. For developers and researchers, it offers insights into unintended biases and areas of improvement, serving as a tool for improving the performance of the model in downstream tasks \cite{bastings2022will, meng2023locating, li2023inferencetime}. However, the scale of LLMs poses unique challenges to explainability. Larger models with more parameters and extensive training data are harder to interpret. Traditional explanation methods such as SHAP values \cite{lundberg2017unified} become less practical for these large-scale models \cite{zhao2023explainability}. 
    Moreover, a comprehensive understanding of LLM-specific phenomena, including in-context learning \cite{halawi2023overthinking, hendel2023incontext, todd2023function, wang-etal-2023-label}, along with addressing issues such as model hallucinations \cite{Ji_2023, chuang2023dola} and inherent biases \cite{devil2023openreview, an-rudinger-2023-nichelle, schick2021selfdiagnosis}, is vital for ongoing refinement in model design.


    In this survey, we focus on explainability methods for pre-trained Transformer-based LLMs, often termed as \textit{base models}. These models often scale up in training data and have billions of parameters; examples include GPT-2 \cite{radford2019language}, GPT-J \cite{chen2021evaluating}, GPT-3 \cite{brown2020language}, OPT \cite{yordanov-etal-2022-shot}, and LLaMA family \cite{touvron2023llama}. 
    In Section \ref{sec:overview}, we categorize and pose research questions based on our survey.
    Based on this categorization, we review explainability methods in Section \ref{sec:exp}, followed by a discussion in Section \ref{sec:app} on how these insights are leveraged.
    We further discuss the evaluation methods and metrics in Section \ref{sec:eval}. 
    Our goal is to synthesize and critically assess contemporary research, aiming to bridge the gap between understanding and practical application of insights derived from complex language models.

\section{Overview}
    \label{sec:overview}
    The field of LLMs is rapidly advancing, making explainability not only a tool for understanding these complex systems but also essential for their improvement. This section categorizes current explainability approaches, highlights the challenges in ethical and controllable generation, and proposes research questions for future exploration.

\paragraph{Categorization of Methods}
    \begin{figure*}[ht]
        \centering
        \includegraphics[width=0.7\textwidth]{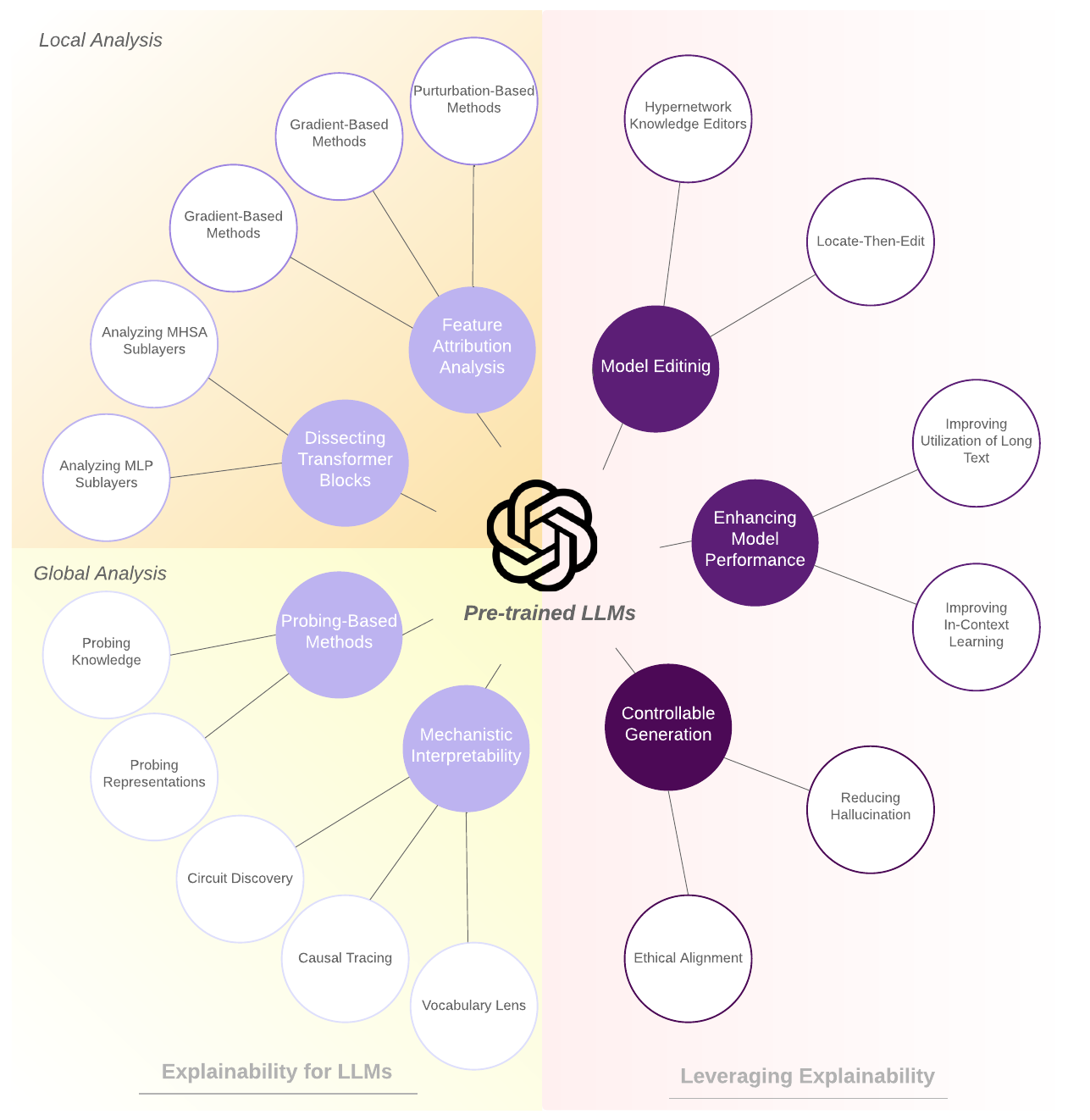} 
        \caption{\small 
        Categorization of literature on explainability in LLMs, focusing on techniques (left, Section \ref{sec:exp}) and their applications (right, Section \ref{sec:app}).
        }
        \label{fig:main}
    \end{figure*}
    We present a structured categorization for the explainability methods and their applications in Figure \ref{fig:main}. 
    Figure \ref{fig:main} presents a structured categorization of explainability methods for pre-trained language models (LMs). We divide these into two broad domains: Local Analysis and Global Analysis. Local Analysis covers feature attribution and transformer block analysis, delving into detailed operations of models. Global Analysis, on the other hand, includes probing-based methods and mechanistic interpretability, offering a comprehensive understanding of model behaviors and capacities. Beyond understanding, we also explore applications of these insights in enhancing LLM capabilities, focusing on model editing, capability enhancement, and controlled generation.

\section{Explainability for Large Language Models}
\label{sec:exp}
\subsection{Local Analysis}
    Local explanations in LLMs aim to elucidate how models generate specific predictions, such as sentiment classification or token predictions, for a given input. This section categorizes local explanation methods into two types: feature attribution analysis and analysis into individual Transformer \cite{Vaswani2017AttentionIA} components.
    
    
\subsubsection{Feature Attribution Explanation} 
    Feature attribution, a local method for explaining a prediction, analysis quantifies the relevance of each input token to a model's prediction. Given an input text $x$ with $n$ tokens $\{x_1, x_2, ..., x_n\}$, a pre-trained language model $f$ outputs $f(x)$. Attribution methods assign a relevance score $R(x_i)$ \cite{modarressi-etal-2022-globenc, ferrando-etal-2022-measuring, modarressi-etal-2023-decompx} to each token $x_i$, reflecting its contribution to $f(x)$. This category includes perturbation-based, gradient-based, and vector-based methods.

\paragraph{Perturbation-Based Methods.}
    Perturbation-based methods, such as LIME \cite{ribeiro2016why} and SHAP \cite{lundberg2017unified}, alter input features to observe changes in model output. 
    However, this removal strategy assumes input features are independent and ignores correlations among them. Additionally, models can be over-confidence even when the predictions are nonsensical or wrong \cite{feng-etal-2018-pathologies}. They also face challenges in efficiency and reliability highlighted in \cite{atanasova-etal-2020-diagnostic}, leading to their diminished emphasis in recent attribution research.
    
\paragraph{Gradient-Based Methods.}
\label{sec:grad}
    One might consider gradient-based explanation methods as a natural approach for feature attribution. This type of method computes per-token importance scores \cite{kindermans2016investigating} using backward gradient vectors. 
    Techniques such as gradient $\times$ input \cite{kindermans2017unreliability} and integrated gradients (IG) \cite{pmlr-v70-sundararajan17a} accumulate the gradients obtained as the input is interpolated between a reference point and the actual input. 
    Despite their widespread use, one main challenge of IG is the computational overhead to achieve high-quality integrals \cite{sikdar-etal-2021-integrated, enguehard2023sequential}
    Their attribution score has also shown to be unreliable in terms of faithfulness \cite{ferrando-etal-2022-measuring} and their ability to elucidate the forward dynamics within hidden states remains constrained. 
    
\paragraph{Vector-Based Methods.}
    Vector-based analyses, which focus on token representation formation, have emerged as a key approach. 
    Approaches range from global attribution from the final output layer to more granular, layer-wise decomposition of token representations \cite{chen-etal-2020-generating-hierarchical, modarressi-etal-2022-globenc}
    Consider decomposing the $i^{th}$ token representation in layer $l \in \{0, 1, 2, ..., L, L+1\}$\footnote{$l=0$ is the input embedding layer and $l = L + 1$ is the language model head over the last decoder layer.}, i.e., $x_i^l \in \{x_1^l, x_2^l, ..., x_N^l\}$, into elemental vectors attributable to each of the $N$ input tokens:
    \begin{equation}
        x_i^l = \sum_{k=1}^{N} x_{i\Leftarrow k}^l
    \end{equation}
    The norm \cite{modarressi-etal-2022-globenc} or the L1 norm \cite{ferrando-etal-2022-measuring} of the attribution vector for the $k^{th}$ input ($x_{i\Leftarrow k}^l$) can be used to quantify its total attribution to $x_i^l$.
    
    Although several established strategies, such as attention rollouts \cite{abnar-zuidema-2020-quantifying, ferrando-etal-2022-measuring, modarressi-etal-2022-globenc}, focus on the global impact of inputs on outputs by aggregating the local behaviors of all layers, they often neglect Feed-Forward Network (FFN) in the analyses due to its nonlinearities. Recent works address this by approximating and decomposing activation functions and constructing decomposed token representations throughout layers \cite{yang-etal-2023-local, modarressi-etal-2023-decompx}. Empirical evaluations demonstrate the efficacy of vector-based analysis and exemplify the potential of such methods in dissecting each hidden state representation within transformers.

\subsubsection{Dissecting Transformer Blocks}
    Tracking Transformer block's \textit{component-by-component internal processing} can provide rich information on its intermediate processing, given the stacked architecture of decoder-based language models \cite{kobayashi2023analyzing}. 
    In a transformer inference pass, the input embeddings are transformed through a sequence of $L$ transformer layers, each composed of a multi-head self-attention (MHSA) sublayer followed by an MLP sublayer \cite{Vaswani2017AttentionIA}. Formally, the representation $x_i^l$ of token $i$ at layer $l$ is obtained by:
    \begin{equation}
        x_i^l = x_i^{l-1} + a_i^l + m_i^l
    \end{equation}
    where $a_i^l$ and $m_i^l$ are the outputs from the $l$-th MHSA and MLP sublayers, respectively \footnote{For brevity, bias terms and layer normalization \cite{ba2016layer} are omitted, as they are nonessential for most of analysis.}. 
    While studies have frequently analyzed individual Transformer components \cite{kobayashi-etal-2020-attention, modarressi-etal-2022-globenc}, the interaction between these sublayers is less explored, presenting an avenue for future research.
    \begin{figure*}[ht]
    \centering
    \subfigure[]{
        \includegraphics[width=0.45\textwidth]{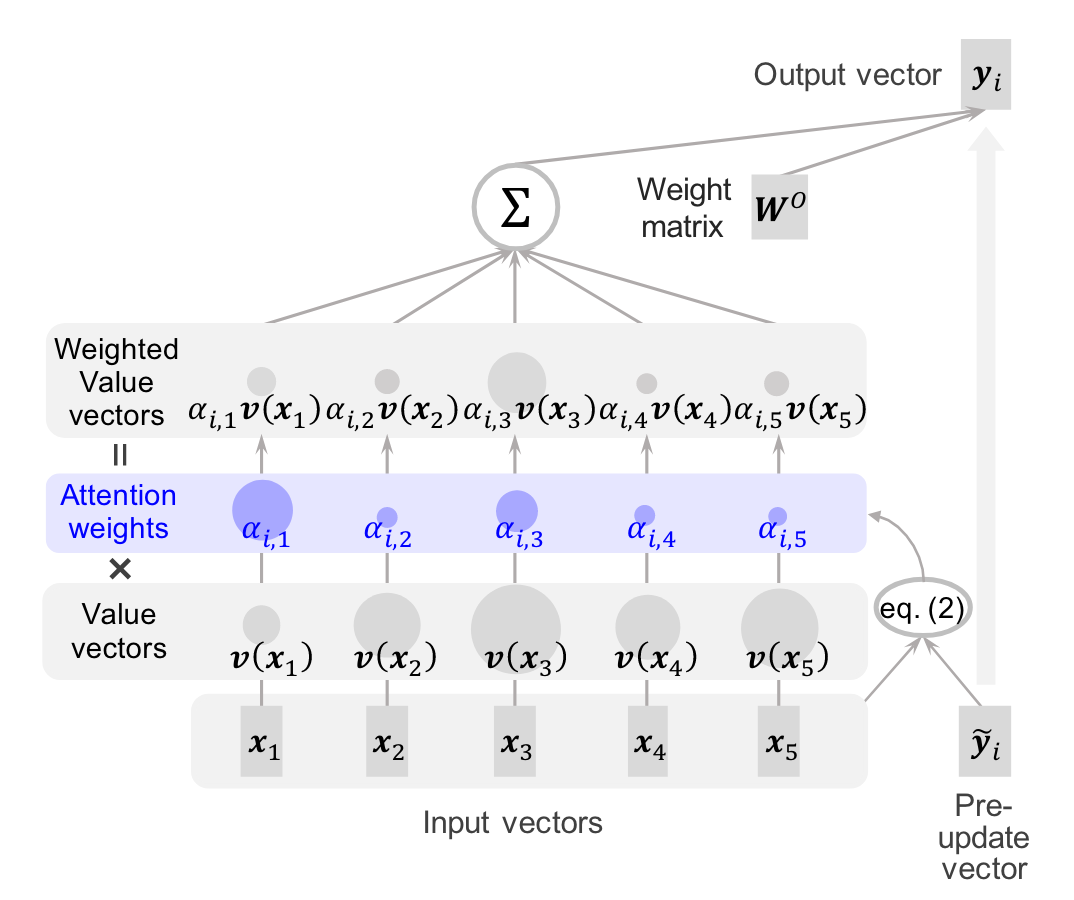}
        \label{fig:mhsa}
    }
    \subfigure[]{
        \includegraphics[width=0.45\textwidth]{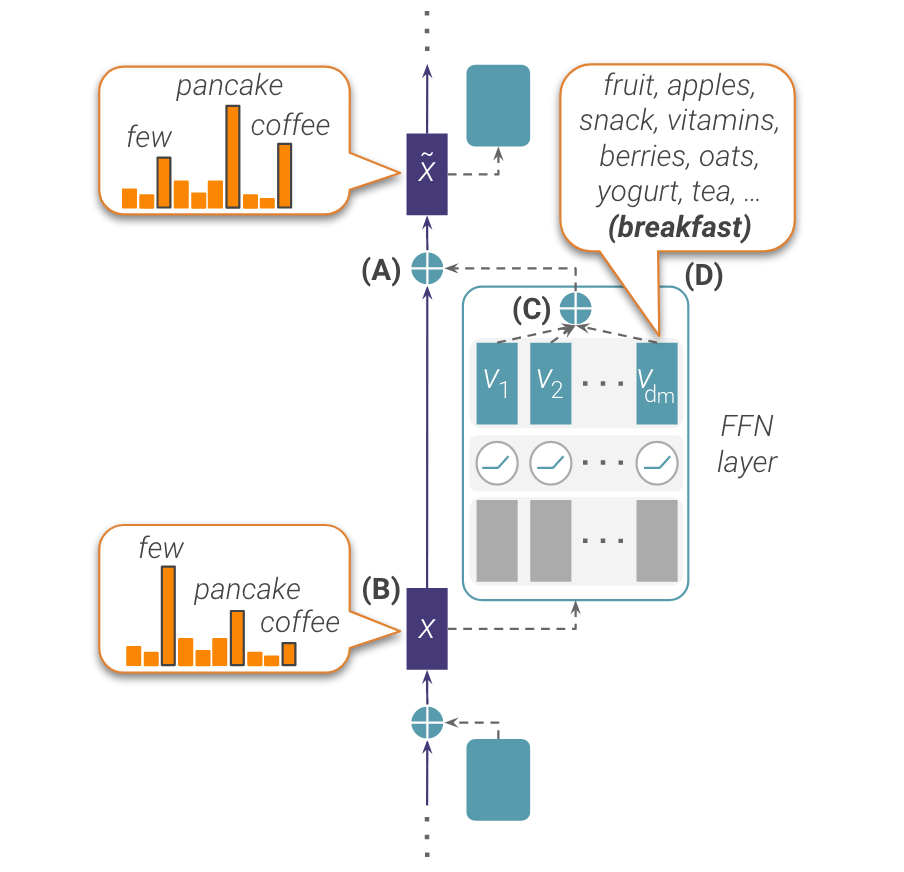}
        \label{fig:mlp}
    }
    \caption{Studied role of each Transformer component. (a) gives an overview of attention mechanism in Transformers. Sizes of the colored circles illustrate the value of the scalar or the norm of the corresponding vector \cite{kobayashi-etal-2020-attention}. (b) analyzes the FFN updates in the vocabulary space, showing that each update can be decomposed to sub-updates corresponding to single FFN parameter vectors, each promoting concepts that are often human-interpretable \cite{geva-etal-2022-transformer}.} 
    \label{fig:block}
\end{figure*}

\paragraph{Analyzing MHSA Sublayers.}
\label{sec:mhsa}
    Attention mechanisms in MHSA sublayers are instrumental in capturing meaningful correlations between intermediate states of input that can explain the model's predictions. 
    Visualizing attention weights and utilizing gradient attribution scores are two primary methods for analyzing these sublayers \cite{zhao2023explainability}. 
    Many studies have analyzed the linguistic capabilities of Transformers by tracking attention weights. \cite{abnar-zuidema-2020-quantifying, katz2023interpreting, kobayashi2023analyzing}. 
    For instance, attention mechanisms typically prioritize specific tokens while diminishing the emphasis on frequent words or special tokens, a phenomenon observable through norm-based analysis metrics, as illustrated in Figure \ref{fig:mhsa} \cite{kobayashi-etal-2020-attention}.  
    In the gradient analysis, some methods calculate gradients as partial derivatives of model outputs with respect to attention weights \cite{10.1145/3459637.3482126}, while others use integrated gradients, which are cumulative versions of these partial derivatives \cite{hao2021selfattention}. Generally, these combined approaches, which integrate attention metrics with gradient information, tend to outperform methods using either metric in isolation. 
    
\paragraph{Analyzing MLP Sublayers.}
\label{sec:mlp}
    More recently, a surge of works have investigated the knowledge captured by the FFN layers \cite{geva-etal-2022-transformer, dai-etal-2022-knowledge}. These layers, consuming the majority of each layer's parameter budget at $8d^2$ compared to $4d^2$ for self-attention layers (where $d$ represents the model's hidden dimension), function akin to key-value memories \cite{geva-etal-2021-transformer}. Here, each "key" is associated with specific textual patterns identified during training, and each "value" generates a corresponding output vocabulary distribution \cite{geva-etal-2021-transformer}. 
    Figure \ref{fig:mlp} focuses in the FFN outputs,  illustrating how each update within these layers can be broken down into sub-updates linked to individual parameter vectors, often encoding concepts that are interpretable to humans \cite{geva-etal-2022-transformer}.
    Additionally, there is an emerging interest in input-independent methods, which interpret model parameters directly, thus eliminating the need for a forward pass \cite{dar-etal-2023-analyzing}.

\subsection{Global Analysis}
    In contrast to local analysis, which focus on elucidating individual model predictions, global analysis aims to understand and explain the knowledge or linguistic properties encoded in the hidden state activations of a model. 
    This section explores two primary approaches to global analysis: probing methods that scrutinize model representations and mechanistic interpretability \cite{transformerCircuits}, an emerging perspective that seeks to reverse engineer the inner workings of deep neural networks.
    
\subsubsection{Probing-Based Method}
    Self-supervised pre-training endows models with extensive linguistic knowledge, derived from large-scale training datasets. Probing-based methods are employed to capture the internal representations within these networks. This approach involves training a classifier, known as a probe, on the network's activations to distinguish between various types of inputs or outputs. In the following sections, we will discuss studies related to probing, categorized based on their objectives, whether it be probing for semantic knowledge or analyzing learned representations.
    
\paragraph{Probing Knowledge.}
    LLMs trained on extensive text corpora, are recognized for their ability to encapsulate context-independent semantic and factual knowledge accessible via textual prompts \cite{petroni-etal-2019-language}. Research in this area primarily focuses on formulating textual queries to extract various types of background knowledge from language models \cite{hewitt-manning-2019-structural, peng2022copen}. Interestingly, probes can sometimes unearth factual information even in scenarios where language models may not reliably produce truthful outputs \cite{hernandez2023inspecting}.

\paragraph{Probing Representations.}
\label{sec:probe}
    LLMs are adept at developing context-dependent knowledge representations. To analyze these, probing classifiers are applied, typically involving a shallow classifier trained on the activations of attention heads to predict specific features. A notable study in this area involved training linear classifiers to identify a select group of attention heads that exhibit high linear probing accuracy for truthfulness \cite{li2023inferencetime}. This research revealed a pattern of specialization across attention heads, with the representation of  ``truthfulness'' predominantly processed in the early to middle layers, and only a few heads in each layer showing standout performance. Such insights pave the way for exploring more complex representations. For instance, research by \cite{li2023emergent} has revealed nonlinear internal representations, such as board game states, in models that initially lack explicit knowledge of the game or its rules.
    

\subsubsection{Mechanistic Interpretability}
\label{sec:mech}
    Mechanistic interpretability seeks to comprehend language models by examining individual neurons and their interconnections, often conceptualized as circuits \cite{transformerCircuits, zhao2023explainability}. This field encompasses various approaches, which can be primarily categorized into three groups: circuit discovery, causal tracing, and vocabulary lens. Each of these approaches offers distinct perspectives and insights into the mechanisms of language models.
    
\paragraph{Circuit Discovery.}
    The circuit-based mechanistic interpretability approach aims to align learned model representations with known ground truths, initially by reverse-engineering the model's algorithm to fully comprehend its feature set \cite{chughtai2023toy}. A prominent example of this approach is the analysis of GPT-2 small \cite{radford2019language}, where a study identified a human-understandable subgraph within the computational graph responsible for performing the indirect object identification (IOI) task \cite{wang2022interpretability}. In IOI, sentences like ``When Mary and John went to the store, John gave a drink'' are expected to be completed with ``Mary''. The study discovered a circuit comprising 26 attention heads -- just 1.1\% of the total (head, token position) pairs -- that predominantly manages this task. This circuits-based mechanistic view provides opportunities to scale our understanding to both larger models and more complex tasks, including recent explorations into In-Context Learning (ICL) \cite{halawi2023overthinking, hendel2023incontext, todd2023function, wang-etal-2023-label}.
    
\begin{figure*}[ht]
        \centering
        \includegraphics[width=1\textwidth]{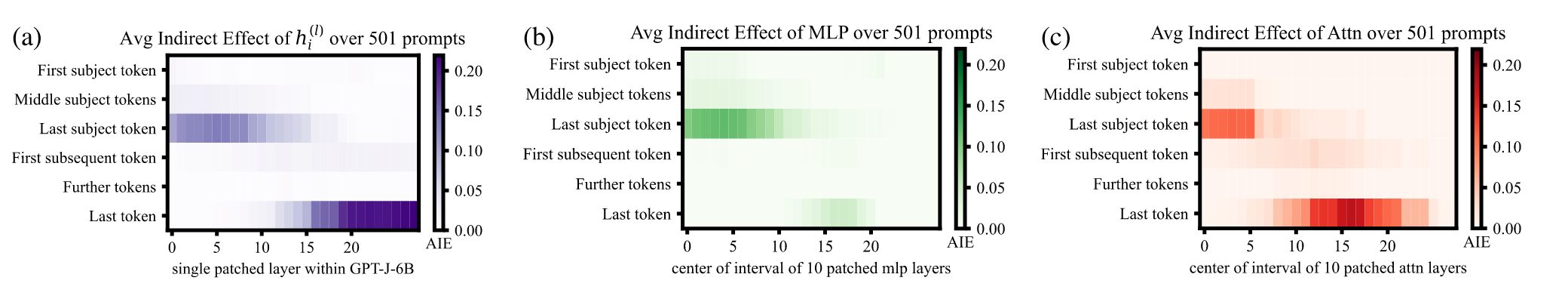} 
        \caption{\small 
        The intensity of each grid cell represents the average causal indirect effect of a hidden state on expressing a factual association. Darker cells indicate stronger causal mediators. It was found that the MLPs at the last subject token and the attention modules at the last token play crucial roles. \cite{meng2023locating}
        }
        \label{fig:causal}
    \end{figure*}

    
\paragraph{Causal Tracing.}
\label{sec:causal}
    The concept of causal analysis in machine learning has evolved from early methods that delineate dependencies between hidden variables using causal graphs \cite{pearl2000models} to more recent approaches like causal mediation analysis \cite{vig2020causal}. This newer method quantifies the impact of intermediate activations in neural networks on their output \cite{meng2023locating}. Specifically, \cite{meng2023locating} assesses each activation's contribution to accurate factual predictions through three distinct operational phases: a \textbf{clean} run generating correct predictions, a \textbf{corrupted} run where predictions are impaired, and a \textbf{corrupted-with-restoration} run that evaluates the ability of a single state to rectify the prediction \cite{meng2023locating}. Termed as \textit{causal tracing}, this approach has identified crucial causal states predominantly in the middle layers, particularly at the last subject position where MLP contributions are most significant (Figure \ref{fig:causal}). This finding underscores the role of middle layer MLPs in factual recall within LLMs.
    

\paragraph{Vocabulary Lens.}
\label{sec:proj}
    Recent work has suggested that model knowledge and knowledge retrieval may be localized within small parts of a language model \cite{geva-etal-2021-transformer} by projecting weights and hidden states onto their vocabulary space. To analyze the components in vocabulary space, we read from each token component $x_{k}^l$ at layer $l$ at the last token position $N$ ($N$ is omitted here), by projecting with the unembedding matrix $E$:
    \begin{equation}
        p_{k}^l = \text{softmax}(E~\text{ln}(x_{k}^l))
    \end{equation}
    where \text{ln} stands for layer normalization before the LM head. \cite{belrose2023eliciting} refines model predictions at each transformer layer and decodes hidden states into vocabulary distributions based on this method. Exploring this avenue further, \cite{geva-etal-2022-transformer} illuminated the role of transformer feed-forward layers in predictions, spotlighting specific conceptual emphases via FFN sub-updates. There is also a growing interest in input-independent methodologies, where model parameters are interpreted directly, bypassing a forward pass \cite{dar-etal-2023-analyzing}.
    
    Augmenting projection-focused interpretations, \cite{din2023jump} first unveiled a feasible application for such projections, suggesting early exit strategies by treating hidden state representations as final outputs. \cite{geva2023dissecting} pinpointed two critical junctures where information propagates to the final predictions via projections and attention edge intervention. While much of the focus has been on how hidden states relate to model outputs, recent works have also highlighted the roles of individual tokens, revealing that their contributions through attention outputs are laden with rich semantic information \cite{ram-etal-2023-token, katz2023interpreting}.

\section{Leveraging Explainability}
\label{sec:app}
    In this section, we discuss how explainability can be used as a tool to debug and improve models. Although various approaches aim to improve model capabilities with fine-tuning or re-training, we focus on methods specifically designed with a strong foundation in model explainability.
    
\subsection{Model Editing}
    Despite the ability to train proficient LLMs, the methodology for ensuring their relevance and rectifying errors remains elusive. In recent years, there has been a surge in techniques for editing LLMs. The goal is to efficiently modify the knowledge or behavior of LLMs within specific domains without adversely affecting their performance on other inputs \cite{yao2023editing}.
    
\paragraph{Hypernetwork Knowledge Editors.}
    
    This type of knowledge editors includes \textit{memory-based} model and editors with \textit{additional parameters}. Memory-based models store all edit examples explicitly in memory based on the explainability finding of key-value memories inside the FFN (Section \ref{sec:mlp}). 
    They can then employ a retriever to extract the most relevant edit facts for each new input, guiding the model to generate the edited fact. 
    SERAC \cite{mitchell2022memorybased}, for instance, adopts a distinct counterfactual model while leaving the original model unchanged.
    Editors with additional parameters introduce extra trainable parameters within LLMs. These parameters are trained on a modified  dataset while the original model parameters remain static. For example, T-Patcher \cite{huang2023transformerpatcher} integrates one neuron (patch) for one mistake in the last layer of the FFN of the model, which takes effect only when encountering its corresponding mistake.

\paragraph{Locate-Then-Edit.}
    The locate-then-edit paradigm first identifies the parameters corresponding to the specific knowledge and then modifies them by directly updating the target parameters. The Knowledge Neuron (KN) method \cite{dai-etal-2022-knowledge} introduces a knowledge attribution technique to pinpoint the ``knowledge neuron'' (a key-value pair in the FFN matrix) that embodies the knowledge and then updates these neurons. ROME \cite{meng2023locating} and MEMIT \cite{meng2023massediting} apply causal tracing (Section \ref{sec:causal}) to locate the editing area. Instead of modifying the knowledge neurons in the FFN, ROME alters the entire matrix. Based on these two methods, PMET \cite{Li2023PMETPM} involves the attention value to achieve better performance.

\subsection{Enhancing Model Capability}
    While LLMs demonstrate versatility in various NLP tasks, insights from explainability can significantly enhance these capabilities.
    This section highlights two key tasks where explainability has shown considerable impact in recent work: improving the utilization of long text \cite{xiao2023efficient, liu2023lost, pope2022efficiently} and enhancing the performance of In-Context Learning (ICL) \cite{hendel2023incontext, halawi2023overthinking, wang-etal-2023-label}.
    
\subsubsection{Improving Utilization of Long Text}
    \begin{figure*}[ht]
        \centering
        \includegraphics[width=1\textwidth]{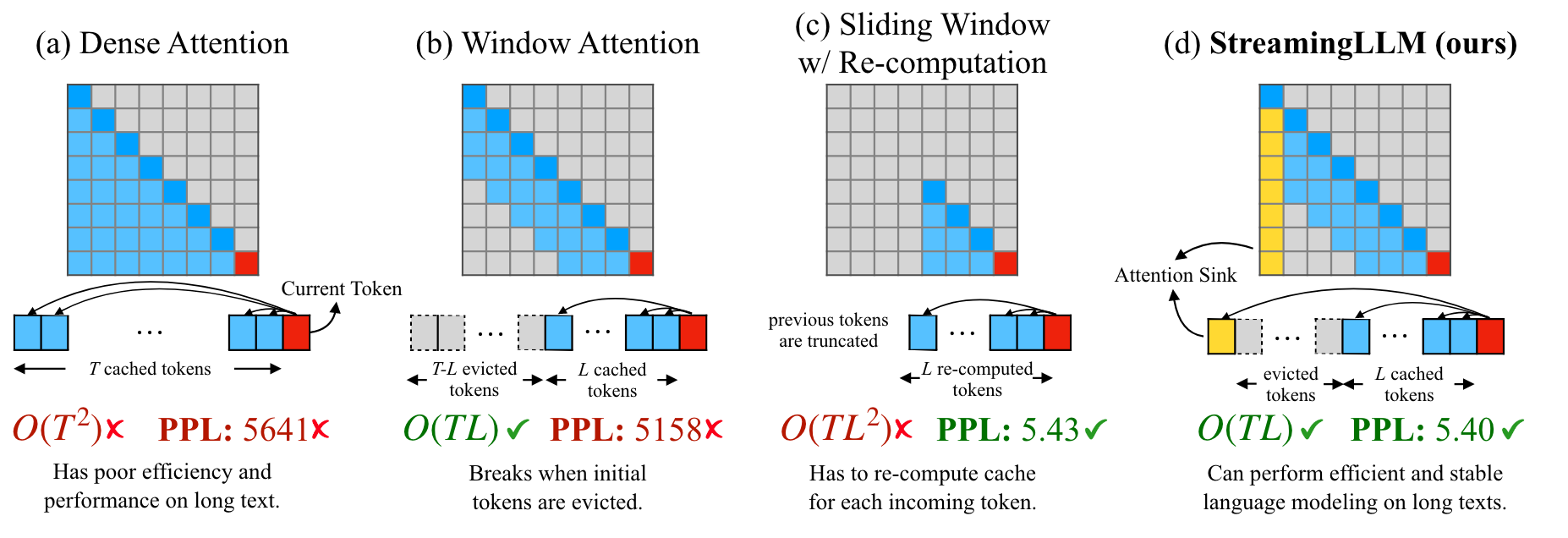} 
        \caption{\small (a) Dense Attention \cite{Vaswani2017AttentionIA} has $O(T^2)$ time complexity and an increasing cache size. Its performance decreases when the text length exceeds the pre-training text length. (b) Window Attention \cite{} caches the most recent $L$ tokens' KV. While efficient in inference, performance declines sharply once the starting tokens' keys and values are evicted. (c) Sliding Window \cite{pope2022efficiently} with Re-computation performs well on long texts, but its $O(TL^2)$ complexity, stemming from quadratic attention in context re-computation, makes it considerably slow. (d) StreamingLLM keeps \cite{xiao2023efficient} the attention sink (several initial tokens) for stable attention computation, combined with the recent tokens. It's efficient and offers stable performance on extended texts.
        }
        \label{fig:streaming}
    \end{figure*}
    The optimization of handling long text aims to enhance the ability of LLMs to capture and effectively utilize content within longer contexts. This is particularly challenging because LLMs tend to struggle with generalizing to sequence lengths longer than what they were pretrained on, such as the 4K limit for Llama-2 \cite{touvron2023llama}.
    \cite{beltagy2020longformer}, maintains a fixed-size sliding window on the key-value (KV) states of the most recent tokens. While this approach ensures constant memory usage and decoding speed after the cache is initially filled, it faces limitations when the sequence length exceeds the cache size \cite{liu2023lost}.
    An innovative solution proposed by \cite{xiao2023efficient} takes advantage of the MHSA explanations (Section \ref{sec:mhsa}) in LLMs, which allocates a significant amount of attention to the initial tokens. They introduce StreamingLLM, a simple and efficient framework that allows LLMs to handle unlimited text without fine-tuning. This is achieved by retaining the "attention sink," which consists of several initial tokens, in the KV states (Figure \ref{fig:streaming}). The authors also demonstrate that pre-training models with a dedicated sink token can further improve streaming performance.

\subsubsection{Improving In-Context Learning}
    In-context Learning (ICL) has emerged as a powerful capability alongside the development of scaled-up LLMs \cite{brown2020language}. 
    ICL stands out because it doesn't require extensive updates to the vast number of model parameters and relies on human-understandable natural language instructions \cite{dong2023survey}. As a result, it offers a promising approach to harness the full potential of LLMs.
    With mechanistic interpretability (Section \ref{sec:mech}), \cite{wang-etal-2023-label} reveal that label words in the demonstration examples function as anchors, which can be used to improve ICL performance with simple anchor re-weighting method. \cite{halawi2023overthinking} study harmful imitation in ICL through vocabulary lens to inspect a model's internal representations (Section \ref{sec:proj}), and identify two related phenomena: \textit{overthinking} and \textit{false induction heads}, the heads in late layers that attend to and copy false information from previous demonstrations, and whose ablation improves ICL performance. Furthermore, using causal tracing (Section \ref{sec:causal}), \cite{hendel2023incontext, todd2023function} find that a small number attention heads transport a compact representation of the demonstrated task, which they call a  \textit{task vector} or \textit{function vector} (FV). These FVs can be summed to create vectors that trigger new complex tasks and improve performance for few-shot prompting \cite{todd2023function}.
 

\subsection{Controllable Generation}
    Though large language models have obtained superior performance in text generation, they sometimes fall short of producing factual content. Leveraging explainability provides opportunities for building inference-time and fast techniques to improve generation models' factuality, calibration, and controllability and align more with human preference.

\subsubsection{Reducing Hallucination}
    Hallucinations in LLMs refer to generated content not based on training data or facts, various factors such as imperfect learning and decoding contribute to this \cite{Ji_2023}. To mitigate hallucinations, initial approaches used reinforcement learning from human feeback \cite{ouyang2022training} and distillation into smaller models such as Alpaca \cite{li2023alpaca}. Leveraging explainability provides a significantly less expensive way to reduce hallucination, enjoying the advantage of being adjustable and minimally invasive. For example, \cite{li2023inferencetime} use as few as 40 samples to locate and find ``truthful'' heads and directions through a trained probe (Section \ref{sec:probe}). They propose inference-time intervention (ITI), a computationally inexpensive strategy to intervene on the attention head to shift the activations in the ``truthful'' direction, which achieves comparable or better performance toward the instruction-finetuned model.


\subsubsection{Ethical Alignment}
    As research on AI fairness gains increasing importance, there have been efforts to detect social bias \cite{fleisig-etal-2023-fairprism, an-rudinger-2023-nichelle} and suppress toxicity \cite{gehman-etal-2020-realtoxicityprompts, schick2021selfdiagnosis} in LMs. Many previous debiasing methods \cite{qian-etal-2022-perturbation} have focused on constructing anti-stereotypical datasets and then either retraining the LM from scratch or conducting fine-tuning. This line of debiasing approaches, although effective, comes with high costs for data construction and model retraining. Moreover, it faces the challenge of catastrophic forgetting if fine-tuning is performed \cite{zhao2023explainability}. While few work has focused on the interpretability of the fairness research, 
    \cite{devil2023openreview} explore interpreting and mitigating social biases in LLMs by introducing the concept of \textit{social bias neurons}. Inspired by the gradient-based attribution method IG (Section \ref{sec:grad}), they introduce an interpretable technique, denoted as intergrated gap gradient (IG\textsuperscript{2}), to pinpoint social bias neurons by back-propagating and integrating the gradients of the logits gap for a selected pair of demographics \footnote{Demographic include properties like gender, sexuality, occupation, etc. 9 common demographics are collected and pairs of demographics are selected to reveal the fairness gap \cite{devil2023openreview}.} Taking this interpretation, they suppress the activations of the pinpointed neurons to mitigate bias. 
    Extensive experiments have verified the effectiveness of this method and have yielded the potential applicability of the explainability method for ethical alignment research in LLMs.

\section{Evaluation}
\label{sec:eval}
    Recently, LLMs such as GPT-4 \cite{openai2023gpt4} have shown impressive abilities to generate natural language explanations for their predictions. However, it remains unclear whether these explanations actually help humans understand the reasoning of the model \cite{zhao2023explainability}. Specifically designed evaluation methods are needed to better assess the performance of explainability methods, such as attribution. Furthermore, calibrated datasets and metrics are required to evaluate the application of explainability to downstream tasks, such as truthfulness evaluation \footnote{Due to space limit, we only discusse the most commonly used evaluation approaches in explainability research}.
    
\subsection{Evaluating Explanation Plausibility}
    One common technique to evaluate the plausibility of attribution analysis is to remove K\% of tokens with the highest or lowest estimated importance to observe its impact on the model output \cite{chen-etal-2020-generating-hierarchical, modarressi-etal-2023-decompx}.
    Another approach to assessing explanation plausibility involves indirect methods, such as measuring the performance of model editing, particularly for ``locate-then-edit'' editing methods, which heavily rely on interpretation accuracy. Recent research \cite{yao2023editing, zhao2023explainability} suggests that having evaluation datasets is crucial for evaluating factual editing in LLMs. Two commonly used datasets for this purpose are ZsRE \cite{levy-etal-2017-zero}, a Question Answering (QA) dataset that employs question rephrasings generated through back-translation, and CounterFact \cite{meng2023locating}, a more challenging dataset that includes counterfacts starting with low scores compared to correct facts.
    
\subsection{Evaluating Truthfulness}
    Model truthfulness is an important metric for measuring the trustworthiness of generative models.
    We expect model outputs to be both informative and factually correct and faithful.
    Ideally, human annotators would label model answers as true or false, given a gold standard answer, but this is often costly.
    \cite{lin2022truthfulqa} propose the use of two fine-tuned GPT-3-13B models (GPT-judge) to classify each answer as true or false and informative or not. 
    Evaluation using GPT-judge is a standard practice on TruthfulQA benchmark, a widely used dataset adversarially constructed to measure whether a language model is truthful in generating answers \cite{askell2021general, li2023inferencetime, chuang2023dola}. The main metric of TruthfulQA is \textbf{true*informative}, a product of scalar truthful and informative scores. This metric not only captures how many questions are answered truthfully but also prevents the model from indiscriminately replying with ``I have no comment'' by assessing the informativeness of each answer.


\section{Conclusion}
    In this survey, we have presented a comprehensive overview of explainability for LLMs and their applications. We have summarized methods for local and global analysis based on the objectives of explanations. In addition, we have discussed the use of explanations to enhance models and the evaluation of these methods. Major future research directions to understanding LLM include developing explanation methods tailored to different language models and making LLMs more trustworthy and aligned with human values by using explainability knowledge. As LLMs continue to advance, explainability will become incredibly vital to ensure that these models are transparent, fair, and beneficial. We hope that this review of the literature provides a useful overview of this emerging research area and highlights open problems and directions for future research.

\bibliography{anthology}



\end{document}